\documentclass[letterpaper]{article} 
\usepackage{aaai25}  
\usepackage{times}  
\usepackage{amsmath}
\usepackage{helvet}  
\usepackage{courier}  
\usepackage[hyphens]{url}  
\usepackage{graphicx} 
\usepackage{natbib}  
\usepackage{caption} 
\usepackage{booktabs}
\usepackage{algorithm}
\usepackage{algorithmicx}
\usepackage{algpseudocode}
\usepackage{newfloat}
\usepackage{listings}
\usepackage{multirow}
\DeclareCaptionStyle{ruled}{labelfont=normalfont,labelsep=colon,strut=off} 
\floatstyle{ruled}
\newfloat{listing}{tb}{lst}{}
\floatname{listing}{Listing}
\title{DSGram: Dynamic Weighting Sub-Metrics for Grammatical Error Correction in the Era of Large Language Models}
\author{
    Jinxiang Xie\textsuperscript{\rm 1}\textsuperscript{\rm 2}\thanks{Work performed as a research intern at Peking University.},
    Yilin Li\textsuperscript{\rm 1},
    Xunjian Yin\textsuperscript{\rm 1},
    Xiaojun Wan\textsuperscript{\rm 1}
}
\affiliations{
    \textsuperscript{\rm 1}Wangxuan Institute of Computer Technology, Peking University,
    
    \textsuperscript{\rm 2}Beijing Jiaotong University

    xiejinxiang@bjtu.edu.cn, wanxiaojun@pku.edu.cn
}

\begin{document}

\maketitle

\begin{abstract}
Evaluating the performance of Grammatical Error Correction (GEC) models has become increasingly challenging, as large language model (LLM)-based GEC systems often produce corrections that diverge from provided gold references. This discrepancy undermines the reliability of traditional reference-based evaluation metrics. In this study, we propose a novel evaluation framework for GEC models, DSGram, integrating Semantic Coherence, Edit Level, and Fluency, and utilizing a dynamic weighting mechanism. Our framework employs the Analytic Hierarchy Process (AHP) in conjunction with large language models to ascertain the relative importance of various evaluation criteria. Additionally, we develop a dataset incorporating human annotations and LLM-simulated sentences to validate our algorithms and fine-tune more cost-effective models. Experimental results indicate that our proposed approach enhances the effectiveness of GEC model evaluations.
\end{abstract}

%
\begin{links}
    \link{Code}{https://github.com/jxtse/GEC-Metrics-DSGram}
    \link{Datasets}{https://huggingface.co/datasets/jxtse/DSGram}
\end{links}

\section{Introduction}

Grammatical Error Correction (GEC) models aims to automatically correct grammatical errors in natural language texts, enhancing the quality and accuracy of written content. Traditionally, the evaluation of GEC models has employed a variety of metrics, categorized into those requiring a reference (reference-based evaluation) and those that do not (reference-free evaluation). 
 
 Reference-based metrics such as BLEU \cite{papineniBleuMethodAutomatic2002}, ERRANT \cite{bryantAutomaticAnnotationEvaluation2017}, and M² \cite{dahlmeierBetterEvaluationGrammatical2012} compare the model-generated text with a correct reference text to evaluate the accuracy of grammatical corrections, and they are widely used in this area. Despite the usefulness of these metrics, they possess inherent limitations. For example, the golden reference may not encompass all potential corrections \cite{choshenInherentBiasesReferencebased2018}, and the alignment of existing automatic evaluation metrics with human judgment is often weak \cite{coyneAnalyzingPerformanceGPT32023}. Additionally, LLMs-based GEC models may excessively correct sentences, resulting in unnecessary editing not captured by traditional metrics \cite{fangChatGPTHighlyFluent2023}.

 \begin{figure}[t!]
    \centering
    \includegraphics[width=1\columnwidth]{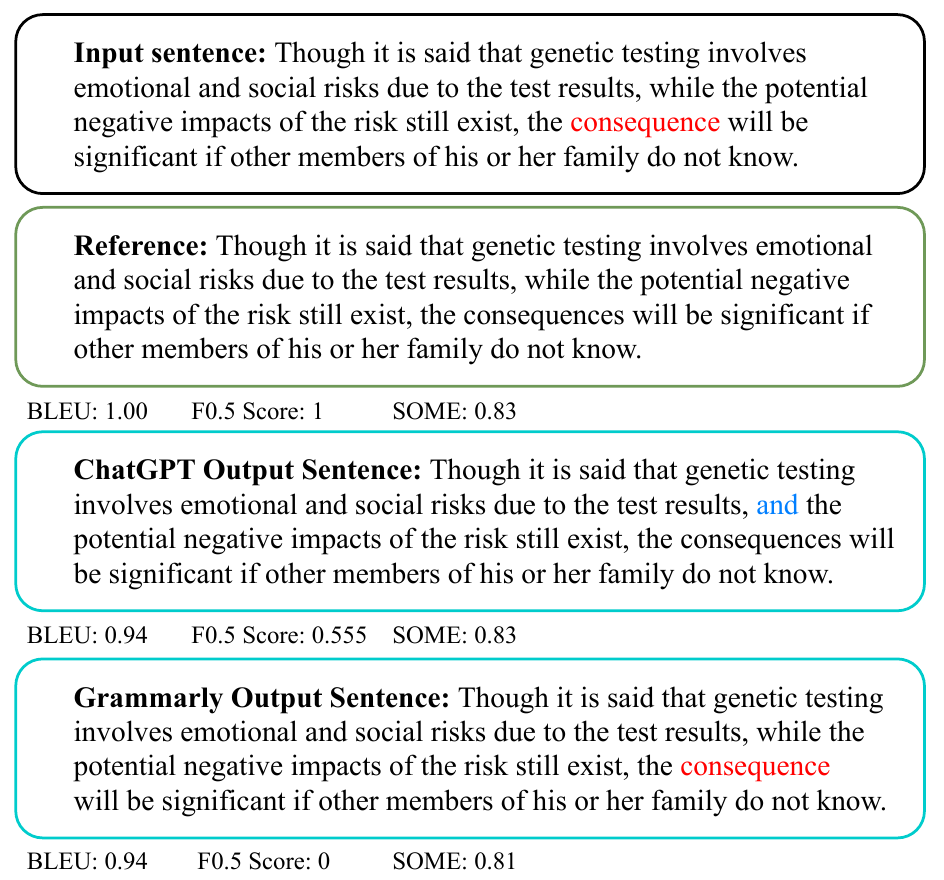}
    \caption{Running examples and evaluation results of several existing metrics. The sentence is from \citet{fangChatGPTHighlyFluent2023}. This figure illustration presents the outcomes of an input sentence processed through two representative GEC models and also the corresponding reference. The metrics' scores are placed under the output. The highlighted in blue represents the over-correction. The highlighted in red indicates poor fluency. Notably, BLEU fails to differentiate between over- and under-correction, whereas SOME cannot capture over-correction.}
    \label{fig:intro}
\end{figure}

 \begin{figure*}[h!]
    \centering
    \includegraphics[width=1\textwidth]{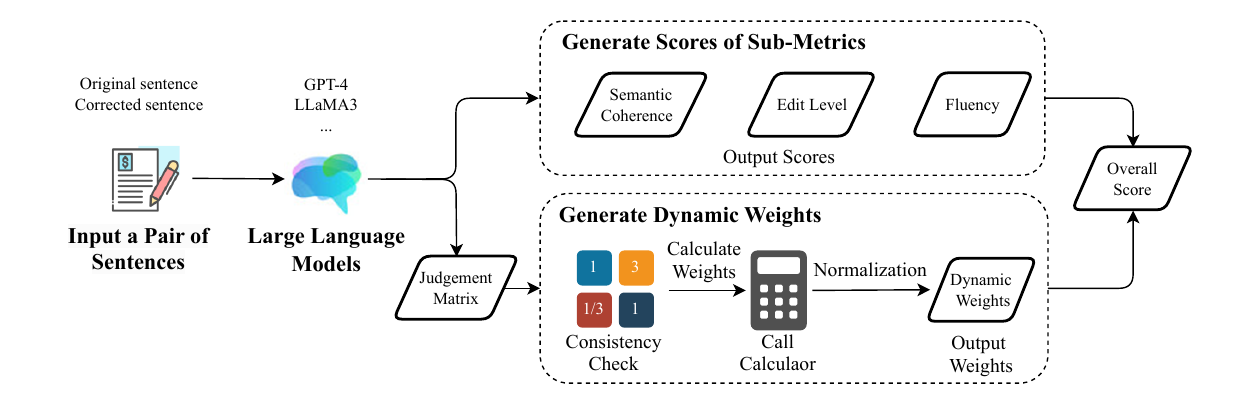} 
    \caption{Architecture of the DSGram method. It begins with the input of sentence pairs (original and corrected). Large language models (such as GPT-4 and LLaMA3) are employed to generate dynamic weights, which are further refined through a judgment matrix and a consistency check. These dynamic weights are normalized and then used to score sub-metrics. The final output consists of both the calculated weights and the overall score for the evaluated correction.}
    \label{fig:flowchart}
\end{figure*}
 
 Conversely, reference-free metrics like Perplexity, GLEU \cite{napolesGroundTruthGrammatical2015}, and SOME \cite{yoshimuraReferencelessSubMetricsOptimized2020} assess the quality of generated text directly, proving beneficial when a correct reference text is unavailable, or when a single (or several) reference is insufficient for GEC evaluation in the era of LLMs. While existing reference-free evaluation metrics like SOME offer an analytical foundation, they no longer fully encompass the scope of GEC evaluation. It is necessary to design new sub-metrics based on the original sub-metrics and emerging requirements. Figure \ref{fig:intro} shows running examples of representative GEC systems and the limitations of current evaluation metrics.

This study seeks to overcome these limitations by introducing a novel evaluation framework, DSGram, for GEC models. DSGram integrates Semantic Coherence, Edit Level, and Fluency, and determines the appropriate weights for these criteria. By adjusting these weights based on the evaluation scenario's context, a more nuanced and context-sensitive evaluation approach can be developed.

Our contributions are summarized as follows: 
\begin{itemize}
    \item We introduce new sub-metrics for GEC evaluation, diverging from previous categorical approaches. Our metrics optimize past sub-metrics by adding an evaluation of over-editing.

    \item  We propose a novel dynamic weighting-based GEC evaluation method, DSGram, which integrates the Analytic Hierarchy Process \citep{AnalyticHierarchyProcess1987} with large language models to ascertain the relative importance of different evaluation criteria.

    \item We present two datasets: DSGram-Eval, created through human scoring, and DSGram-LLMs, a larger dataset designed to simulate human scoring for fine-tuning. Both datasets utilize sentences from the CoNLL-2014 and BEA-2019 test sets to facilitate rigorous evaluation.

\end{itemize}

\section{Related Work}

Numerous studies have focused on evaluating GEC models. This section provides an overview of key research utilizing LLMs for GEC assessment.

\subsection{Model-Based Evaluation Metrics}
Model-based evaluation metrics have garnered significant attention, especially in the domain of GEC. BLEURT \cite{sellamBLEURTLearningRobust2020} is a versatile metric that evaluates based on (prediction, reference) pairs. It employs a scalable pretraining phase where it learns to predict automatically generated signals of supervision from semantically comparable synthetic pairs.

A burgeoning trend in automatic evaluation involves the direct application of LLMs for assessment purposes. \citet{liuGEvalNLGEvaluation2023} have LLMs generate numerical ratings by interpreting descriptions of evaluation criteria through a Chain-of-Thought method \cite{weiChainofThoughtPromptingElicits2023}. \citet{sottanaEvaluationMetricsEra2023} demonstrate the viability of GPT-4 in GEC assessment, highlighting an approach that uses natural language instructions to define evaluation criteria.

\subsection{Reference-Free Evaluation}
Reference-free evaluation is a method of assessing the performance of models without relying on reference. \citet{asanoReferencebasedMetricsCan2017} integrate three sub-metrics, which are Grammaticality, Fluency, and Meaning Preservation, to surpass reference-based metrics. They employ a language model and edit-level metrics as Fluency and Meaning Preservation sub-metrics, respectively, although these sub-metrics are not tailored for manual evaluation. The final score is determined through a weighted linear summation of each individual evaluation score.

SOME is a reference-free GEC metric that follows the approach of \citet{asanoReferencebasedMetricsCan2017}. It utilizes three distinct BERT models, each dedicates to one scoring aspect. The researchers construct a novel dataset for training these BERT models by annotating the outputs from various GEC systems on the CoNLL-2013 test set across the three scoring dimensions. 

However, in real-world applications, the use of just three evaluation metrics presents a limitation, as it hinders the generation of a holistic score in an inherently intuitive way. Consequently, this constraint diminishes its effectiveness in directing the training of GEC models. In the sphere of GEC evaluation, \citet{wuRethinkingMaskedLanguage2023} assessed ChatGPT and observed a tendency towards over-correction, which might be attributed to its extensive generative capacities as a large language model. Similarly, \citet{fangChatGPTHighlyFluent2023} noted that ChatGPT makes over-corrections, leading to revised sentences with high fluency. These findings align with the outcomes of our own experiments, which indicate that existing reference-free metrics do not adequately reflect these issues.

\subsection{GEC Dataset with Human Scoring}
There are very few GEC datasets with human evaluation scores. The dataset annotated by \citet{yoshimuraReferencelessSubMetricsOptimized2020} includes individual scores for Grammaticality, Fluency, and Meaning Preservation, but lacks an overall score. GJG15 \citep{grundkiewiczHumanEvaluationGrammatical2015} and SEEDA \cite{kobayashiRevisitingMetaevaluationGrammatical2024a} are manually annotated but provide ranking information rather than scores. \citet{sottanaEvaluationMetricsEra2023} conducted ratings based on the gold standard, which presents certain limitations.

\citet{sottanaEvaluationMetricsEra2023} suggested the potential of GPT-4 as a reviewer, prompting us to consider using GPT-4 to annotate a dataset that simulates human scoring for GEC model evaluation.

\section{Dynamic Weighting Sub-Metrics for GEC}
DSGram comprises two main components: score generation and weight generation. By applying specific weights to the generated scores, an overall score is obtained. Figure \ref{fig:flowchart} illustrates the method's flowchart.

\subsection{Generating Scores}
\subsubsection{Sub-Metrics Definition}
\begin{figure}[t!]
    \centering
    \includegraphics[width=0.45\textwidth]{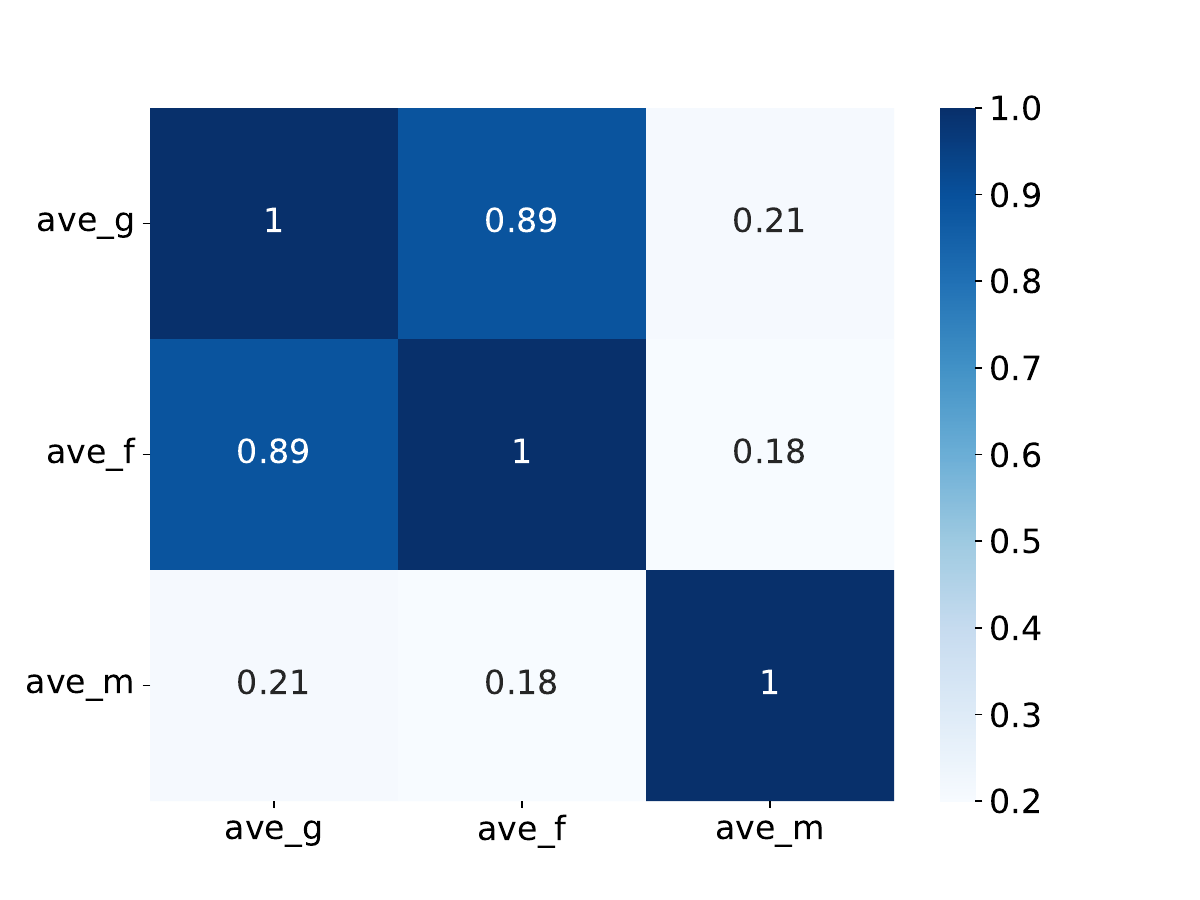} 
    \caption{Heat map of the correlations among the three sub-metrics of SOME. avg\_g, avg\_f, and avg\_m denote the average scores for Grammaticality, Fluency, and Meaning Preservation, respectively. The correlation between avg\_g and avg\_f is notably strong at 0.89.} 
    \label{fig:heatmap1}
\end{figure}

Upon our analysis, the three sub-metrics introduced by \citet{asanoReferencebasedMetricsCan2017}, exhibit redundancy and insufficiency. We compute the correlation of these metrics using the SOME's dataset, and the results are depicted in Figure \ref{fig:heatmap1}.

\begin{figure}[t!]
    \centering
    \includegraphics[width=0.45\textwidth]{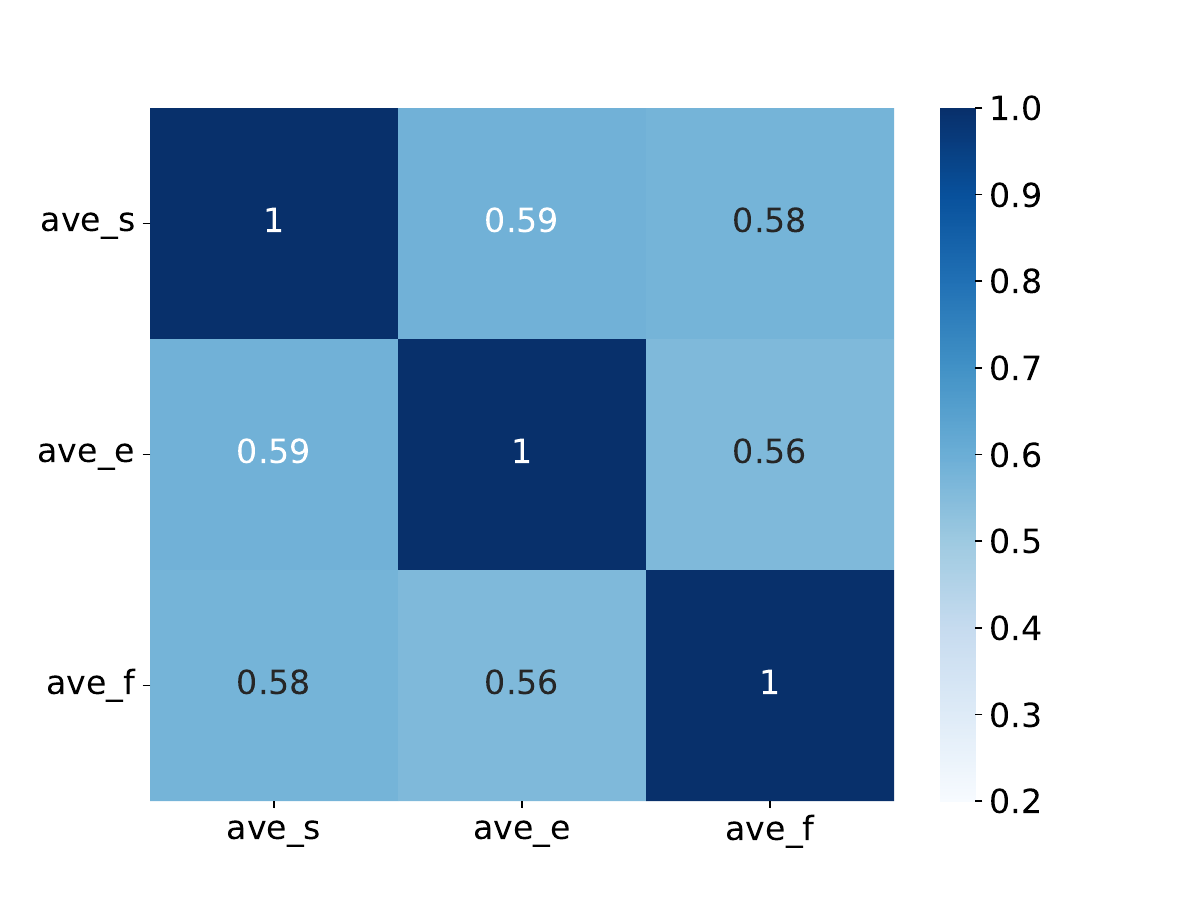} 
    \caption{Heat map of our three sub-metrics correlations. avg\_s, avg\_e, and avg\_f denote the average scores for Semantic Coherence, Edit Level and Fluency, respectively. The correlation has become more evenly distributed.}
    \label{fig:heatmap2}
\end{figure}

The heatmap reveals a high correlation between Grammaticality and Fluency. Hence, we have combined these two metrics into a single "Fluency" measure. Furthermore, to address the issue of over-corrections present in LLM-based GEC models, we have incorporated a new sub-metric called "Edit Level" to evaluate concerns associated with excessive corrections. Based on the computation of 200 human-annotated scores, The correlation of these novel sub-metrics is shown in Figure \ref{fig:heatmap2}. It can be observed that our classification criteria have improved the sub-metrics' distribution.

In our definition, the meanings of the three sub-metrics are as follows:

\paragraph{Semantic Coherence: } The degree to which the meaning of the original sentence is preserved in the corrected sentence. It evaluates whether the corrected sentence conveys the same intended meaning as the original, without introducing semantic errors or altering the core message.

\paragraph{Edit Level: } The extent to which the GEC model has modified the sentence. It assesses whether the corrections made are necessary and appropriate, or if the sentence has been unnecessarily or excessively altered, deviating from the original prose more than required.

\paragraph{Fluency: } The grammatical correctness and the natural flow of the corrected sentence. It evaluates whether the sentence adheres to proper grammar rules, has a coherent structure, and reads smoothly without awkward phrasing or unnatural constructions.

These three sub-metrics comprehensively cover the key aspects that GEC models need to consider in the era of large language models.

\subsubsection{Sub-Metrics Score Generation}

\begin{figure*}[h]
    \centering
    \includegraphics[scale=0.5]{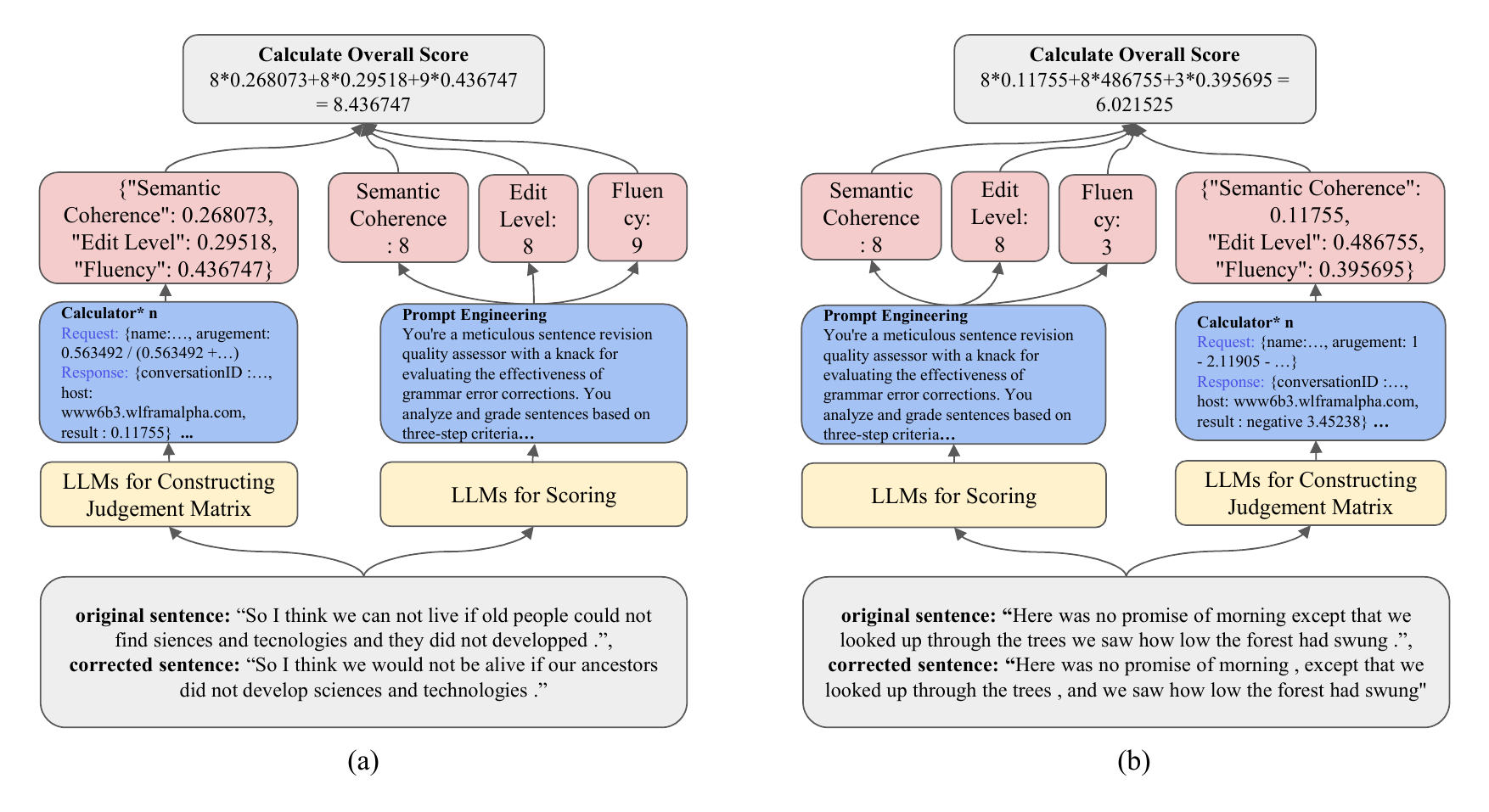} 
    \caption{The DSGram score computation processes for two different sentences. Sentence (a) represents a casual daily dialogue where the emphasis is on Fluency. Sentence (b) is a more formal expression, thus placing a greater emphasis on the Edit Level.}
    \label{fig:example}
\end{figure*}

We employ prompt engineering techniques such as Chain-of-Thought \cite{weiChainofThoughtPromptingElicits2023}, few-shots \cite{brownLanguageModelsAre2020} and output reason before scoring \cite{chuBetterLLMEvaluator2024} to craft the prompts, enabling LLMs to generate scores based on the aforementioned three sub-metrics. The specific prompts can be found in Appendix.

The method is applicable to a variety of different LLMs. Our experiments reveal that the GPT-4 and GPT-4o exhibit greater scoring consistency and alignment with human judgments. Models such as LLaMA2 and LLaMA3 struggle to adhere to structured prompts and do not effectively grasp the scoring tasks. To reduce evaluation costs and enhance model usability, we annotate a dataset simulating human scoring with GPT-4 to fine-tune open-source LLMs, exploring their feasibility for scoring.
We utilize GPT-4 to annotate DSGram-LLMs and subsequently fine-tune the LLaMA2-13B and LLaMA3-8B models on this dataset, thereby confirming their consistency with human scores.

In summary, we achieve the automatic generation of evaluation scores with prompting GPT-4 and open-source LLMs like LLaMA.

\subsection{Generating Dynamic Weights}
The prevalent approach using sub-metrics methods typically involves assigning specific weights to compute an overall score. However, we contend that in the context of GEC evaluation tasks, human judges may have varying considerations for different sentences. It is essential to adjust the weights according to the evaluation scenario; for formal documents such as legal texts and medical instructions, a stricter standard is required for Semantic Coherence and Edit Level. In contrast, for more relaxed contexts like dialogues, Fluency should be given priority in the assessment.

We employ LLMs in conjunction with the Analytic Hierarchy Process (AHP) to generate varying weights for different sentences, thereby making the weighted scores more aligned with human scoring. AHP is a decision-making framework that decomposes complex decisions into a hierarchy of simpler, more manageable elements through pairwise comparisons. It synthesizes subjective judgments into objective priorities that are appropriately weighted based on a structured evaluation of criteria and objectives. 

The weight generation algorithm involves the following key steps:

\noindent\textbf{Constructing the Judgement Matrix: } For each criterion, utilize LLMs to construct a pairwise comparison matrix $A = (a_{ij})$, where $a_{ij}$ represents the importance of criterion $i$ relative to criterion $j$. The specific prompt for this process is documented in the Appendix.

\noindent\textbf{Consistency Check: } Compute the consistency index (CI) and consistency ratio (CR) to check the consistency of the pairwise comparison matrix.
$$\text{CI} = \frac{\lambda_{\max} - n}{n - 1}, \text{CR} = \frac{\text{CI}}{\text{RI}},$$
where RI is the Random Index for a matrix of order $n$. For instance, an RI value of 0.58 is used for a 3×3 matrix, while for a 4×4 matrix, RI is 0.90, and this pattern continues. A CR below 0.1 indicates that the pairwise comparison matrix possesses sufficient consistency.

\noindent\textbf{Calculating the Eigenvector and Eigenvalue: } Solve the eigenvalue problem $A \mathbf{w} = \lambda_{\max} \mathbf{w}$ to obtain the maximum eigenvalue $\lambda_{\max}$ and the corresponding eigenvector $\mathbf{w}$, which is then normalized.

\noindent\textbf{Calculating the Composite Weight: } Compute the composite weights by multiplying the weights at each level.

\begin{algorithm*}
\caption{Dynamic Weight Calculation for GEC Evaluation using AHP and LLMs}
\begin{algorithmic}[1]
\Procedure{DynamicWeightCalculation}{$C = \{C_1, C_2, \ldots, C_n\}$}

\Comment{Input: Criteria set (e.g., Semantic Coherence, Edit Level, Fluency)}
\ForAll{pair of criteria $(C_i, C_j)$, each sentence $S_k$}
\Comment{\textbf{Construct Judgment Matrices:}}
    \State Construct judgment matrix $A = [a_{ij}]$ utilizing LLMs for $S_k$
    \Comment{$a_{ij}$ represents relative importance of $C_i$ to $C_j$}
\EndFor

\ForAll{judgment matrix $A$}
\Comment{\textbf{Consistency Check:}}
    \State Calculate Consistency Index (CI) and Consistency Ratio (CR)
    \If{$CR \geq \theta$}
        \State Adjust ratings in $A$ until $CR < \theta$
    \EndIf
\EndFor

\ForAll{judgment matrix $A$}
\Comment{\textbf{Calculate Weights:}}
    \State Compute principal eigenvector $\mathbf{w} = (w_1, w_2, \ldots, w_n)$ of $A$
    \State Normalize to obtain weights $w_i$ for each criterion $C_i$
\EndFor

\ForAll{sentence $S_k$}
\Comment{\textbf{Compute Aggregated Scores:}}
    \State Calculate weighted score $W_k = \sum_{i=1}^{n} w_i \cdot s_{k,i}$
    \Comment{$s_{k,i}$ is the score of $S_k$ on criterion $C_i$}
\EndFor

\State \textbf{Return} aggregated scores for each $S_k$
\EndProcedure
\end{algorithmic}
\label{alg:1}
\end{algorithm*}

The procedure for this approach is outlined in Algorithm \ref{alg:1}. Figure \ref{fig:example} depicts an illustration of this method for two distinct sentence pairs. Sentence (a) represents a casual daily dialogue where the emphasis is on Fluency. On the other hand, sentence (b) is a more formal expression with a higher level of professionalism, thus placing a greater emphasis on the Edit Level.

\section{Experiments and Analysis}

\subsection{Datasets Preparation}
We mainly evaluate our metrics using the SEEDA dataset \cite{kobayashiRevisitingMetaevaluationGrammatical2024a}, and compare them with existing metrics. The SEEDA corpus comprises human-rated annotations at two distinct evaluation levels: edit-based (SEEDA-E) and sentence-based (SEEDA-S), and encompasses corrections from 12 cutting-edge neural systems, including large language models, as well as corrections made by humans.

In addition to SEEDA, we also manually annotate an evaluation dataset with human scores for validation purposes. We name this human-annotated dataset DSGram-Eval. For details on annotating this dataset, please refer to the Appendix.

Furthermore, we construct an additional training set by utilizing system outputs from the BEA-2019 shared task official website\footnote{\url{https://www.cl.cam.ac.uk/research/nl/bea2019st/}}. After removing biased output groups, a sample of approximately 2500 entries is randomly selected. Then GPT-4 is employed to generate the sub-metrics scores, which we refer to as DSGram-LLMs. This set is subsequently used for model training.

\subsection{Meta-Evaluation of Our Metrics}

We evaluate the performance of various GEC models using the DSGram metric on the SEEDA dataset. GPT-3.5 secures the highest score in our DSGram metric, aligning with the findings in \citet{fangChatGPTHighlyFluent2023}, which report ChatGPT as having the highest human score in fluency. Despite it exhibiting substandard scores in metrics like $M^2$, we still regard it as an outstanding GEC system. Our metric demonstrates equally outstanding performance on models like T5 \cite{rotheSimpleRecipeMultilingual2021} and TransGEC \cite{fangTransGECImprovingGrammatical2023} that excel in GLEU and ERRANT. However, the REF-F model, despite boasting the highest SOME score, underperforms significantly when assessed using our metric. The results are shown in Table \ref{baseline1}.

\begin{table*}[h]
  \centering
  \begin{tabular}{l|ccccc}
    \hline
    GEC Model & \textbf{$M^2$} & \textbf{$ERRANT$} & \textbf{$GLEU$}  & \textbf{$SOME$} &  \textbf{$DSGram$} \\
    \hline
    BERT-fuse \cite{kiyonoEmpiricalStudyIncorporating2019} & 62.77 & 58.99 & 68.5 & 0.8151 &  9.3853 \\
    GECToR-BERT \cite{omelianchukGECToRGrammaticalError2020} & 61.83 & 58.05 & 66.56 & 0.8016 & 9.1473 \\
    GPT-3.5 & 53.5 & 44.12 & 65.93 & 0.8379 & \textbf{9.6310}  \\
    PIE \cite{awasthiParallelIterativeEdit2019} & 59.93 & 55.89 & 67.83 & 0.8066 &  9.0342 \\
    REF-F (most fluency references by experts) & 47.48 & 33.24 & 60.34 & \textbf{0.8463} & 9.0534 \\
    REF-M (minimal edit references by experts)& 60.12 & 54.77 & 67.27 & 0.8155 & 9.4661 \\
    Riken-Tohoku \cite{kanekoEncoderDecoderModelsCan2020} & 64.74 & 61.88 & 68.37 & 0.8123 & 9.4442 \\
    T5 \cite{fangTransGECImprovingGrammatical2023} & 65.07 & 60.65 & 68.81 & 0.8202 & 9.4983   \\
    TransGEC \cite{fangTransGECImprovingGrammatical2023} & \textbf{68.08} & \textbf{64.43} & \textbf{70.20} & 0.8200 & 9.5711 \\
    
    \hline
  \end{tabular}
  \caption{\label{baseline1}
    Scores of common GEC models w.r.t various existing metrics and DSGram on SEEDA dataset. The bold part indicates the model with the highest score under each metric. Certain GEC models with low scores have been omitted from this table.
  }
\end{table*}

We calculate the correlation between the ranking according to DSGram and human ranking on the SEEDA dataset and compare it with existing metrics. Specifically, we convert the concrete data obtained in Table \ref{baseline1} into human rankings and compare them with the human rankings in the dataset, obtaining the corresponding correlation. The comparison is presented in Table \ref{tab:humanrank} and the obtained correlation is presented in Table \ref{baseline2}.

The results indicate that the correlation of our metric with human feedback surpasses that of all conventional reference-based metrics, as well as reference-free metrics like GLEU, Scribendi Score and DSGram weighted by 0.33.

\begin{table*}[]
    \centering
    \scalebox{1}{}{
    \begin{tabular}{l c | l c}
    \hline
    GEC Model & DSGram Score & GEC Model & Human Score\\
    \hline
    GPT-3.5 & 9.631 & REF-F & 0.992 \\
    TransGEC & 9.571	& GPT-3.5 & 0.743 \\
    T5 & 9.498 & T5 & 0.179 \\
    REF-M & 9.466	& TransGEC & 0.175\\
    Riken-Tohoku & 9.444 & REF-M & 0.067 \\
    UEDIN-MS & 9.411 & BERT-fuse & 0.023 \\
    BERT-fuse & 9.385 & Riken-Tohoku & -0.001 \\
    GECToR-BERT & 9.147 & PIE & -0.034 \\
    REF-F & 9.055 & LM-Critic	& -0.163 \\
    PIE	& 9.034 & TemplateGEC	& -0.168 \\
    GECToR-ens & 9.030 & GECToR-BERT & -0.178 \\
    LM-Critic & 9.017 & UEDIN-MS & -0.179 \\ 
    BART & 8.934 & GECToR-ens	& -0.234 \\
    TemplateGEC	& 8.899 & BART & -0.300 \\
    INPUT & 8.127 & INPUT	& -0.992 \\
    \hline
    \end{tabular}
    }
    \caption{A comparative analysis of the DSGram Score and the Human Score on SEEDA dataset, as ranked by the DSGram Score, indicates a favorable correlation, with our scores exhibiting a close alignment to those assigned by humans.}
    \label{tab:humanrank}
\end{table*}

\begin{table*}[ht]
\centering
\scalebox{1}[1]{
\begin{tabular}{l|cc|cc|cc|cc}
\hline
\multirow{2}{*}{Metrics} & \multicolumn{4}{c|}{System-level} & \multicolumn{4}{c}{Sentence-level} \\
 & \multicolumn{2}{c|}{SEEDA-S} & \multicolumn{2}{c|}{SEEDA-E} & \multicolumn{2}{c|}{SEEDA-S} & \multicolumn{2}{c}{SEEDA-E} \\
& r & $\rho$ & r & $\rho$ & Acc & $\tau$ & Acc & $\tau$\\
\hline
 $M^2$ & 0.658 & 0.487 & 0.791 & 0.764 & 0.512 & 0.200 & 0.582 & 0.328 \\
 $ERRANT$ & 0.557 & 0.406 & 0.697 & 0.671 & 0.498 & 0.189 & 0.573 & 0.310 \\
\hline
 $GLEU$ & 0.847 & 0.886 & 0.911 & 0.897 & 0.673 & 0.351 & 0.695 & 0.404 \\
 $SOME$ & 0.892 & 0.867 & 0.901 & \textbf{0.951} & 0.768 & \textbf{0.555} & 0.747 & \textbf{0.512} \\
 $IMPARA$ \cite{maeda-etal-2022-impara} & \textbf{0.911} & 0.874 & 0.889 & 0.944 & 0.761 & 0.540 & 0.742 & 0.502 \\
 $Avg\_DSGram (G.w)$ & 0.797 & 0.790 & 0.922 & 0.930 & 0.648 & 0.296 & 0.661 & 0.323 \\
 $DSGram (L.w)$ & 0.880 & 0.853 & 0.906 & 0.895 & 0.631 & 0.263 & 0.649 & 0.298 \\ 
 $DSGram (G.w)$ & 0.880 & \textbf{0.909} & \textbf{0.927} & 0.944 & \textbf{0.776} & 0.551 & \textbf{0.750} & 0.499 \\
\hline
\end{tabular}
}
\caption{
System-level and sentence-level meta-evaluation results of common GEC models. We follow the \citet{kobayashiRevisitingMetaevaluationGrammatical2024a}, use Pearson (r) and Spearman ($\rho$) for system-level and Accuracy (Acc) and Kendall ($\tau$) for sentence-level meta-evaluations. G.w denotes GPT-4 weighted, and L.w denotes LLaMA3-70B weighted models.
The sentence-based human evaluation dataset is denoted SEEDA-S and the edit-based one is denoted SEEDA-E. 
The score in bold represents the metrics with the highest correlation at each granularity.
}
\label{baseline2}
\end{table*}

\subsection{Validation of LLMs Scoring}
To verify the broad applicability of our metrics, we conduct tests on DSGram-Eval dataset by using various LLMs (including few-shot and finetuned LLaMA models) to generate the sub-metric scores and then calculate the correlation between the sub-metric/overall scores with human scores. The correlation results in Table \ref{llm_corr} indicate that GPT-4 consistently shows a high correlation with human scores across three sub-metrics. In contrast, LLaMA3-70B is found to be highly correlated in Semantic Coherence and Fluency but less so in Edit Level. Since the few-shot LLaMA performs poorly, we fine-tune LLaMA on the DSGram-LLMs dataset. The fine-tuned LLaMA3-8B and LLaMA2-13B models outperform their original few-shot outcomes, demonstrating that LLMs can achieve higher human-aligned scores for GEC tasks, while fine-tuning smaller LLMs serves to reduce evaluation costs and enhance model usability.

\begin{table*}[h!]
  \centering
  \begin{tabular}{l|cccc}
    \hline
    LLMs  & Semantic Coherence & Edit Level & Fluency  & Overall\\
    \hline
    GPT-4 (Zero-shot) & \textbf{0.724} & \textbf{0.839} & \textbf{0.797} & \textbf{0.772} \\
    LLaMA3-70B (Zero-shot)  & 0.399 &  0.349 & 0.574  & 0.607 \\
    LLaMA2-13B (Fine-tuned) & 0.315 & 0.239 & 0.258 & 0.382                         \\
    LLaMA3-8B (Fine-tuned)  & 0.372 & 0.331 &  0.361 & 0.419                       \\
    LLaMA2-13B (5-shots) & 0.215 & 0.189 & 0.245 & 0.306                         \\
    LLaMA3-8B (5-shots)  & 0.327 & 0.243 &  0.261 & 0.373                       \\
    \hline
  \end{tabular}
  \caption{\label{llm_corr}
    Pearson correlation with human scoring on DSGram-Eval across various LLMs
  }
\end{table*}

\subsection{Validation of Dynamic Weights}
To ascertain the efficacy of the dynamic weights in DSGram, we conduct a comparison between the average overall score assigned by human annotators and the score weighted by generated dynamic weights derived from human-labeled sub-metrics. 

We calculate the AHP Human Score by applying dynamic weights to the human-annotated Semantic Coherence, Edit Level, and Fluency scores. The Pearson correlation coefficient between the AHP Human Score and the human-annotated overall score is \textbf{0.8764}. Subsequently, we compute the overall score derived from the average weighting method and correlated it with the human scoring overall score. By assigning an equal weight of 0.33 to each metric, we obtain the AVG Human Score. The calculation reveal a Pearson correlation coefficient of \textbf{0.8544} between the AVG Human Score and the human scoring overall score.

Our experimental findings indicate that when the overall score is adjusted using dynamic weights, it significantly corresponds with the evaluations provided by human annotators, and the dynamic weighting approach outperforms the average weighting method at a relaxed significance level.

It is suggested that, although human annotators do not consciously consider the specific weights of each metric while annotating, they implicitly gauge their relative importance. This intuitive evaluation process resembles the creation of a judgment matrix, and our method effectively mimics the human scoring procedure.

To further verify the effectiveness, we select three distinctly different text datasets: OpenSubtitles\footnote{\url{https://www.opensubtitles.org/}}, Justia (CaseLaw)\footnote{\url{https://law.justia.com/cases/}}, and Wikipedia Corpus, to represent everyday conversations, legal documents, and technical explanations, respectively. For each dataset, we generate weights for three sub-metrics and evaluate the internal consistency of these weights using Cronbach's Alpha coefficient \cite{cronbachCoefficientAlphaInternal1951}. Specifically, we preprocess each dataset to extract sub-metric scores and then calculate Cronbach's Alpha based on the generated weights to assess the consistency and reliability of the sub-metrics across different datasets.

The experimental results demonstrate that the Cronbach's Alpha coefficients for the OpenSubtitles, Justia (CaseLaw), and Wikipedia Corpus datasets are \textbf{0.76}, \textbf{0.82}, and \textbf{0.79}, respectively, all exceeding the acceptable threshold of 0.7. 

This indicates that the weights generated in different scenarios exhibit high consistency and reliability. The weights in the everyday conversation scenario show good consistency, while the weights in the legal and technical documentation scenarios perform even better. Overall, this experiment validates the effectiveness of the weight generation algorithm across different text types, demonstrating its robustness and reliability in diverse applications.

\section{Discussion}

\textbf{Human Feedback}\quad For reference-free evaluation metrics, their effectiveness is often gauged by how closely they mirror human judgments. Although ChatGPT is rated as the best model in terms of grammar by human reviewers, it is rated as the worst or second-worst model in terms of semantics and over-correction \cite{sottanaEvaluationMetricsEra2023}. This suggests that human feedback also seems to have its limitations.

\noindent\textbf{LLMs as Reviewers} \quad The utilization of large language models to construct evaluation metrics has been found to closely approximate human feedback, thereby enhancing the accuracy of assessment. Additionally, the dynamic evaluation approach demonstrates a degree of rationality, which can potentially be transferred to other forms of assessment.

\noindent\textbf{Applications}\quad The metrics in question may prove applicable in guiding the training of GEC models. For large-scale models, the rate of gradient descent during the training process may vary across different tasks. For instance, the model's ability to correct semantic errors may have already saturated, whereas there is still room for improvement in terms of fluency correction. Employing this metric to construct the loss function could potentially enhance the training of GEC models.

\section{Conclusions}

This study presents an evaluation framework for Grammatical Error Correction models that integrates Semantic Coherence, Edit Level, and Fluency through a dynamic weighting system. By leveraging the AHP in conjunction with LLMs, we have developed a method that dynamically adjusts the importance of different evaluation criteria based on the context, leading to a more nuanced and accurate assessment.

Through extensive experiments, our methodology has demonstrated strong alignment with human judgments, particularly when utilizing advanced LLMs like GPT-4. The dynamic weighting system has shown considerable promise in mirroring the intuitive scoring processes of human annotators, thereby validating its application in various contexts.

Looking forward, several avenues for future research are evident. First, it is imperative to conduct more extensive tests of this method across a broader range of LLMs, including Claude, GLM and Qwen. Moreover, a more comprehensive validation of the dynamic evaluation approach is required, to explore its applicability in diverse evaluation contexts.

\section*{Acknowledgements}
This work was supported by Beijing Science and Technology Program (Z231100007423011) and Key Laboratory of Science, Technology and Standard in Press Industry (Key Laboratory of Intelligent Press Media Technology). We appreciate the anonymous reviewers for their helpful comments. Xiaojun Wan is the corresponding author.

\bibliography{aaai25}

\end{document}